\documentclass[10pt,twocolumn,letterpaper]{article}

\usepackage{cvpr}
\usepackage{times}
\usepackage{epsfig}
\usepackage{graphicx}
\usepackage{amsmath}
\usepackage{amssymb}

\usepackage{algorithm,algorithmic}
\usepackage{amsmath}
\usepackage{bbm}
\usepackage{amssymb}
\usepackage{amsthm}
\usepackage{url}
\usepackage{makecell}

\usepackage{multirow}
\usepackage{multicol}
\usepackage{booktabs}
\usepackage{rotating}
\usepackage{threeparttable}

\usepackage{graphicx}
\usepackage{subcaption}

\newtheorem{theorem}{Theorem}
\newtheorem{lemma}{Lemma}

\DeclareMathOperator*{\argmax}{arg\,max}

\newcommand{\xadv}[1]{{\bf x}_{adv}^{(#1)}}

\newcommand{\sourceimage}{{source-image}\xspace}

\newcommand{\targetimage}{{target-image}\xspace}

\newcommand{\advimage}{{adv-image}\xspace}

\newcommand{\queryimage}{{\it query image}\xspace}

\newcommand{\boundaryimage}{{\it boundary-image}\xspace}

\newcommand{\maliciousclass}{{\it malicious-class}\xspace}

\newcommand{\name}{{QEBA}\xspace}

\newcount\Comments  % 1 suppresses notes to selves in text
\usepackage{color}
\definecolor{darkgreen}{rgb}{0,0.5,0}
\definecolor{darkblue}{rgb}{0,0,0.5}
\definecolor{purple}{rgb}{1,0,1}
\newcommand{\kibitz}[2]{\ifnum\Comments=0\textcolor{#1}{#2}\fi}
\newcommand{\Huichen}[1]{\kibitz{blue}      {[HL: #1]}}

\usepackage[pagebackref=true,breaklinks=true,letterpaper=true,colorlinks,bookmarks=false]{hyperref}

\cvprfinalcopy % *** Uncomment this line for the final submission

\def\cvprPaperID{9582} % *** Enter the CVPR Paper ID here

\ifcvprfinal\fi
\begin{document}

%%%%%%%%% TITLE
\title{QEBA: Query-Efficient Boundary-Based Blackbox Attack}
\author{Huichen Li$^{1*}$\quad Xiaojun Xu$^{1}$
\thanks{The first two authors contribute equally. This work was done while they were interns at Ant Financial. To appear at CVPR 2020. The code is available at \url{https://github.com/AI-secure/QEBA}} % last sentences added for arxiv version
\quad Xiaolu Zhang$^{2}$\quad Shuang Yang$^{2}$\quad Bo Li$^{1}$
\\
$^{1}$University of Illinois at Urbana-Champaign\quad $^{2}$ Ant Financial
}

% \and
\maketitle
\thispagestyle{empty}

%%%%%%%%% ABSTRACT
\begin{abstract}
   Machine learning (ML), especially deep neural networks (DNNs) have been widely used in various applications, including several safety-critical ones (e.g. autonomous driving). As a result, recent research about \emph{adversarial examples} has raised great concerns. Such adversarial attacks can be achieved by adding a small magnitude of perturbation to the input to mislead model prediction. While several whitebox attacks have demonstrated their effectiveness, which assume that the attackers have full access to the machine learning models; blackbox attacks are more realistic in practice. 
%   In this paper, we propose a Query-Efficient Boundary-based Blackbox Attack (\name) to estimate model decision boundary given classification output by leveraging the intrinsic dimensionalities of inputs.
In this paper,  we propose a Query-Efficient Boundary-based blackbox Attack (\name) based only on model's final prediction labels.
%   We theoretically show why estimating model gradient is not efficient in terms of blackbox attack, and provide the optimality analysis for our dimension reduced decision boundary estimation.
We theoretically show why previous boundary-based attack with gradient estimation on the whole gradient space is not efficient in terms of query numbers, and provide optimality analysis for our dimension reduction-based gradient estimation.
   On the other hand, we conducted extensive experiments on ImageNet and CelebA datasets to evaluate \name. We show that compared with the state-of-the-art blackbox attacks, \name is able to use a smaller number of queries to achieve a lower magnitude of perturbation with 100\% attack success rate. We also show case studies of attacks on real-world APIs including MEGVII Face++ and Microsoft Azure.
%   \bo{Azure?}.\bo{do I miss any exp results here? if you have numbers want to add, it is also good}
   
\end{abstract}

%%%%%%%%% BODY TEXT
\begin{figure*}[htbp]
    \centering
    \includegraphics[width=\textwidth]{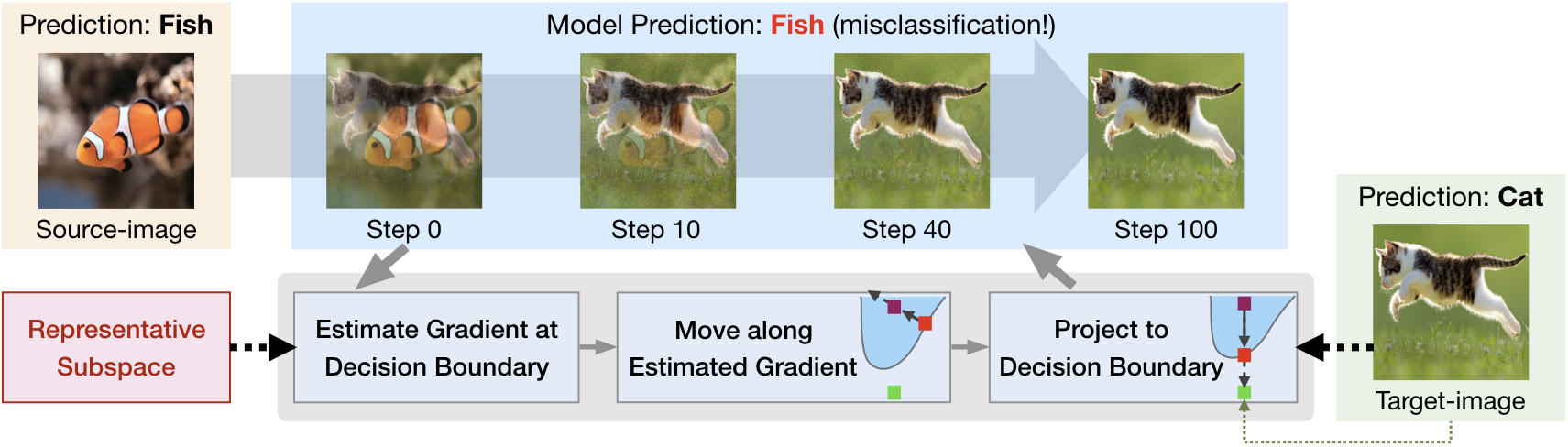}
    \caption{Pipeline of \name. In this example, the attack goal is to obtain an \advimage that looks like a cat (\targetimage) but be misclassified as a fish ($y_{mal}$). We start from a source-image together with an optimized subspace. We then iterativelly perform gradient estimation with queries, move along the estimated direction, and project the new instance to the decision boundary by binary search towards the \targetimage till converge. The grey solid arrows indicate steps within each iteration. 
    In particular, we show a toy example of how the source-image (purple rectangles) is moved towards the \targetimage (green rectangles), while the intermediate projected \boundaryimage is shown as red rectangles.}
    \label{fig:pipeline}
    % \vspace{-0.5cm}
\end{figure*}

\section{Introduction}
% 1. machine learning is vulnerable, many attack
% 2. blackbox attack is more realistic, transfearbility -- it cannot that it works, 
% gradient estimation - finite diffxxx, but they require logit output
% only based on output -- example of APIs that allow ***
% 3. challenges, 1. #queries, 2. how to reduce dimension, litterateurs -- three perspectives are important, so we plan to make a comprehensive studies about these. 
% 4. in the meantime, what's the key factor for such reduction -- mention the intuition about the theorem and rho
% 5.  exp: datasets, and models, APIs, beat all the stoa
% 6. contribution list: 1. propose blackbox from three xxxxx, and we did comprehensive studies for three types of query reduction; 2. from theoretic perspective, xxxx show xxx is important-- much smaller #queries, and lower magnitude of perturbation, 100% attack success; 3. extensive experiments for models/datasets. 4. real-world APIs

Recent developments of machine learning (ML), especially deep neural networks (DNNs), have advanced a number of real-world applications, including object detection~\cite{ren2015faster}, drug discovery~\cite{chen2018rise}, and robotics~\cite{lenz2015deep}. In the meantime, several safety-critical applications have also adopted ML, such as autonomous driving vehicles~\cite{chen2015deepdriving} and surgical robots~\cite{richter2019open,mlsurgicalrobotics}.
However, recent research have shown that machine learning systems are vulnerable to \emph{adversarial examples}, which are inputs with small magnitude of adversarial perturbations added and therefore cause arbitrarily incorrect predictions during test time~\cite{eykholt2017robust,xiao2018generating,carlini2017towards,goodfellow2014explaining,chaowei2018characterizing,chaowei2018spatially}.
Such adversarial attacks have led to great concerns when applying ML to real-world applications. Thus in-depth analysis of the intrinsic properties of these adversarial attacks as well as potential defense strategies are required.

First, such attacks can be categorized into whitebox and blackbox attacks based on the attacker's knowledge about the victim ML model. In general, the whitebox attacks are possible by leveraging the gradient of the model --- methods like fast gradient sign method (FGSM) ~\cite{goodfellow2014explaining}, optimization based attack~\cite{carlini2017towards}, projected gradient descent based method (PGD)~\cite{madry2017towards} have been proposed. However, whitebox attack is less practical, given the fact that most real-world applications will not release the actual model they are using. In addition, these whitebox attacks are shown to be defendable~\cite{madry2017towards}.
As a result, blackbox adversarial attack have caught a lot of attention in these days.
In blackbox attack, based on whether an attacker needs to query the victim ML model, there are query-free (e.g. transferability based attack) and query-based attacks. Though \emph{transferability} based attack does not require query access to the model, it assumes the attacker has access to the large training data to train a substitute model, and there is no guarantee for the attack success rate. The query based attack includes score-based and boundary-based attacks. Score-based attack assumes the attacker has access to the class probabilities of the model, which is less practical compared with boundary-based attack which only requires the final model prediction, while both require large number of queries.

% In this paper, we propose a Query-Efficient Boundary-based blackbox Attack (\name) to estimate model decision boundary given only the classification output by leveraging the intrinsic dimensionalities of inputs. 
In this paper,  we propose Query-Efficient Boundary-based blackbox Attack (\name) based only on model's final prediction labels as a general framework to minimize the query number.
Since the gradient estimation consumes the majority of all the queries, the main challenge of reducing the number of queries for boundary-based blackbox attack is that a high-dimensional data (e.g. an image) would require large number of queries to probe the decision boundary. As a result, we propose to search for a small representative subspace for query generation.
% that can serve as the supports for the original gradient space. \Huichen{supports?}
In particular, queries are generated by adding perturbations to an image. We explore the subspace optimization methods from three novel perspectives for perturbation sampling: 1) spatial, 2) frequency, and 3) intrinsic component.
The first one leverages spatial transformation (e.g. linear interpolation) so that the sampling procedure can take place in a low-dimensional space and then project back to the original space.
The second one uses intuition from image compression literature and samples from low frequency subspace and use discrete consine transformation (DCT)~\cite{guo2018low} to project back.
The final one performs scalable gradient matrix decomposition to select the major principle components via principle component analysis (PCA)~\cite{wold1987principal} as subspace to sample from.
% The resulting boundary-based attack from the three methods are called \name-S, \name-F, and \name-I respectively.
% The first one is the spatial transformed subspace. We first sample a low-dimensional vector and then apply spatial transformation (e.g. linear interpolation) to project it back to the original space to get perturbations. 
% Secondly, we propose the low-frequency subspace enabled blackbox attack (\name-F), which leverage discrete consine transformation (DCT)~\cite{guo2018low} to select the low frequency subspace to perform efficient queries against the victim model.
% Finally we also propose the intrinsic component subspace based blackbox attack (\name-I) by performing scalable gradient matrix decomposition to select the major principle components via principle component analysis (PCA)~\cite{wold1987principal} as subspace to conduct the queries.
In addition
% to the three proposed query efficient \name methods
, we theoretically prove the optimality of them on estimating the gradient compared with estimating the gradient directly over the original space.

To demonstrate the effectiveness of the proposed blackbox attack \name methods, 
% , and analyzing different dimension reduction methods, 
we conduct extensive experiments on high dimensional image data including ImageNet~\cite{deng2009imagenet} and CelebA~\cite{liu2018large}. We perform attacks on the ResNet model~\cite{he2016deep}, and show that compared with the-state-of-the-art blackbox attack methods, the different variations of \name can achieve lower magnitude of perturbation with smaller number of queries (attack success rate 100\%). 
\Huichen{I deleted one sentence here because its expression is not accurate.}
% In particular, the \name-S performs slightly better in terms of achieving lower magnitude of perturbation.
In order to show the real-world impact of the proposed attacks, we also perform \name against online commercial APIs including MEGVII Face++\cite{facepp-compare-api} and Microsoft Azure\cite{azure-detect-api}. Our methods can successfully attack the APIs with perturbations of reasonable magnitude.
Towards these different subspaces, our conjecture is that the over-all performance on different subspaces depends on multiple factors including dataset size, model smoothness, adversarial attack goals etc. Therefore, our goal here is to make the first attempt towards providing sufficient empirical observations for these three subspaces, while further extensive studies are required to compare different factors of these subspaces, as well as identifying new types of subspaces.
The \textbf{contributions} of this work are summarized as follows: 
% 1) We propose a general query-efficient blackbox attack \name to reduce the number of queries based on boundary-based attack;
% 2) We propose three different subspace optimization-based blackbox attack approaches, including spatial transformed attack (\name-S), low frequency attack (\name-F), and intrinsic component based attack (\name-I);
1) We propose a general Query-Efficient Boundary-based blackbox Attack \name to reduce the number of queries based on boundary-based attack. The \name contains three variations based on three different representative subspaces including spatial transformed subspace, low frequency subspace, and intrinsic component subspace;
2) We theoretically demonstrate that gradient estimation in the whole gradient space is inefficient in terms of query numbers, and we prove the optimality analysis for our proposed query-efficient gradient estimation methods;
3) We conduct comprehensive experiments on two high resolution image datasets: ImageNet and CelebA. All the different variations of \name outperform the state-of-the-art baseline method by a large margin;
4) We successfully attack two real-world APIs including Face++\cite{facepp-compare-api} and Azure\cite{azure-detect-api} and showcase the effectiveness of \name.

% \begin{enumerate}
    % \item We propose a general query efficient blackbox attack \name to reduce the number of queries based on boundary-based attack;
    % \item We theoretically demonstrate that only gradient estimation is insufficient for performing blackbox attacks, and we provide the optimality analysis for our proposed query efficient decision boundary estimation;
    % \item We propose three different subspace optimization based blackbox attack approaches, including spatial transformed attack (\name-S), low frequency attack (\name-F), and intrinsic component based attack (\name-I);
    % \item We conduct comprehensive experiments on two high resolution image datasets: ImageNet and CelebA. DIfferent variations of \name outperform the state-of-the-art baselines by a large margin;
    % \item We also successfully attack two real-world APIs including xxxx\bo{fill} and showcase the effectiveness of \name.
% \end{enumerate}

% \section{Background}

\section{Problem Definition}
Consider a $k$-way image classification model $f({\bf x})$ where ${\bf x}\in\mathbb{R}^m$ denotes the input image with dimension $m$, and $f({\bf x})\in\mathbb{R}^k$ represents the vector of confidence scores of the image belonging to each classes. 
In boundary-based black-box attacks, the attacker can only inquire the model with queries $\{ {\bf x}_i \}$ (a series of updated images) and get the predicted labels % $\tilde{y}=F({\bf x}) = \argmax_j [f({\bf x})]_j$
$\tilde{y}_i=F({\bf x_i}) = \argmax_j [f({\bf x_i})]_j,$
where $[f]_j$ represents the score of the $j$-th class. The parameters in the model $f$ and the score vector $\bf s$ are not accessible.

There is a \targetimage ${\bf x}_{tgt}$ with a \emph{benign label} $y_{ben}$. Based on the \emph{malicious label} $y_{mal}$ of their choice, the adversary will start from a \sourceimage ${\bf x}_{src}$ selected from the category with label $y_{mal}$, and move ${\bf x}_{src}$ towards ${\bf x}_{tgt}$ on the pixel space while keeping $y_{mal}$ to guarantee the attack. 
An image that is on the decision boundary between the two classes (e.g. $y_{ben}$ and $y_{mal}$) and is classified as $y_{mal}$ is called \boundaryimage. 

The adversary's goal is to find an \emph{adversarial image}(\advimage) ${\bf x}_{adv}$ such that $F({\bf x}_{adv}) = y_{mal}$ and $D({\bf x}_{tgt}, {\bf x}_{adv})$ is as small as possible, where $D$ is the distance metric (usually $L_2$-norm or $L_\infty$-norm distance). By definition, \advimage is a \boundaryimage with an optimized (minimal) distance from the \targetimage. 
In the paper we focus on targeted attack and the approaches can extend to untargeted scenario naturally.

\section{Query-Efficient Boundary-based blackbox Attack (\name)}
\label{sec:subspaces}
In this section we first introduce the pipeline of \name which is based on HopSkipJumpAttack (HSJA)~\cite{chen2019hopskipjumpattack}. We then illustrate the three proposed query reduction approaches in detail. We provide the theoretic justification of \name in Section~\ref{sec:dimred-theory}.
The pipeline of the proposed Query-Efficient Boundary-based blackbox Attack (QEBA) is shown in Figure \ref{fig:pipeline} as an illustrative example. The goal is to produce an \advimage that looks like ${\bf x}_{tgt}$ (cat) but is mislabeled as the malicious label (fish) by the victim model.
% The \sourceimage is the starting point of the attack instance.
First, the attack initializes the \advimage with ${\bf x}_{src}$.
% Then it performs an iterative algorithm which takes the \targetimage and a chosen representative subspace as input.
Then it performs an iterative algorithm consisting of three steps: \textbf{estimate gradient at decision boundary} which is based on the proposed representative subspace, \textbf{move along estimated gradient}, and \textbf{project to decision boundary} which aims to move towards ${\bf x}_{tgt}$.
% There are two inputs to the attack algorithm: the \targetimage and the optimized representative subspace. The \targetimage is fixed and used during the attack. The optimized representative subspace is generated before the attack and also fixed. The attack itself is an iterative algorithm consisting of three steps: estimating gradient at decision boundary, moving towards estimated gradient, and projecting to decision boundary.

% Before we introduce the three steps in our pipeline in detail, 
% we first 
First, define the adversarial prediction score $S$ and the indicator function $\phi$ as:
\begin{align}
    S_{{\bf x}_{tgt}}({\bf x}) &= [f({\bf x})]_{y_{mal}} - \max_{y \neq y_{mal}} [f({\bf x})]_y,\\
    \phi_{{\bf x}_{tgt}}({\bf x}) &= \text{sign}(S_{{\bf x}_{tgt}}({\bf x})) =
    \begin{cases}
    1 & \text{if } S_{{\bf x}_{tgt}}({\bf x})\geq 0;\\
    -1 & \text{otherwise}.
    \end{cases}
\end{align}
% \begin{align}
%     \phi({\bf x}) &=
%     \begin{cases}
%     1 & \text{if } F(x) \text{ equals } y_{mal} \\
%     -1 & \text{otherwise}.
%     \end{cases}
% \end{align}
We abbreviate the two functions as $S({\bf x})$ and $\phi({\bf x})$ if it does not cause confusion. In boundary-based attack, the attacker is only able to get the value of $\phi$ but not $S$.
% Then $\bf x$ is adversarial if and only if $S({\bf x})\geq 0$, and the adversary can know the value of $\phi$ in the decision-based black-box setting.

In the following, we first introduce the three interative steps in the attack in Section~\ref{sec:iterative_attack}, then introduce three different methods for generating the optimized representative subspace in Section~\ref{sec:name-S}-\ref{sec:name-I}. 

\subsection{General framework of \name}
\label{sec:iterative_attack}

\begin{figure}
    \centering
    \includegraphics[width=\linewidth]{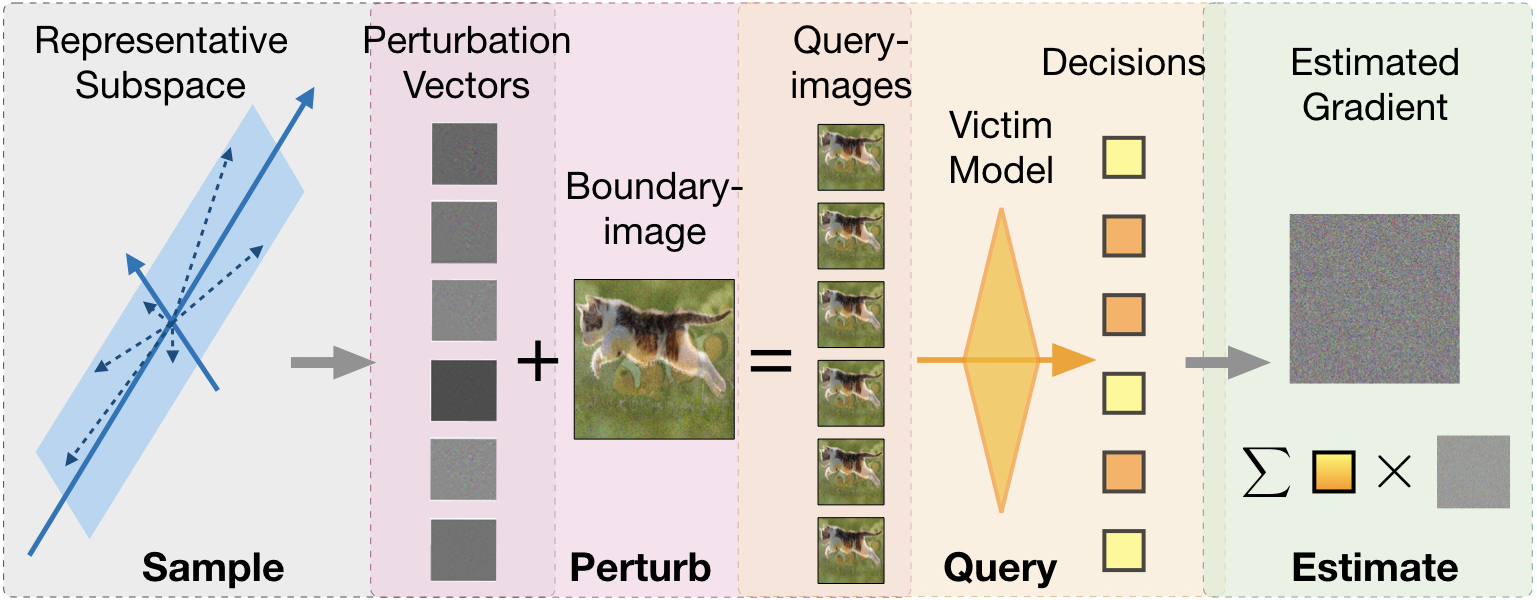}
    \caption{Query model and estimate gradient near the decision boundary.}
    \label{fig:gradient_estimation}
    % \vspace{-0.5cm}
\end{figure}

\paragraph{Estimate gradient at decision boundary}
Denote $\xadv{t}$ as the \advimage generated in the $t$-th step. The intuition in this step is that we can estimate the gradient of $S(\xadv{t})$ using only the access to $\phi$ if $\xadv{t}$ is at the decision boundary. This gradient can be sampled via Monte Carlo method:
% Therefore, if we assume that $\xadv{t}$ is at the decision boundary, then the gradient of $S$ w.r.t. an \advimage which lies on the decision boundary can be estimated via Monte Carlo method:
\begin{align}
    \widetilde{\nabla S} = \frac1B \sum_{i=1}^B \phi(\xadv{t}+\delta {\bf u}_b) {\bf u}_b
    \label{eq:MC_gradient_estimation}
\end{align}
where $\{{\bf u}_b\}$ are $B$ randomly sampled perturbations with unit length and $\delta$ is a small weighting constant. An example of this process is shown in Figure \ref{fig:gradient_estimation}. The key point here is how to sample the perturbation ${\bf u}_b$'s and we propose to draw from a representative subspace in $\mathbb{R}^n$.

Formally speaking, let $W=[w_1, \ldots, w_n] \in \mathbb{R}^{m\times n}$ be $n$ orthonormal basis vectors in $\mathbb{R}^m$, meaning $W^\intercal W = I$. Let $\text{span}(W) \subseteq \mathbb{R}^m$ denote the $n$-dimensional subspace spanned by $w_1, \ldots, w_n$. We would like to sample random perturbations from $\text{span}(W)$ instead of from the original space $\mathbb{R}^m$. In order to do that, we sample ${\bf v}_b \in \mathbb{R}^n$ from unit sphere in $\mathbb{R}^n$ and let ${\bf u}_b = W {\bf v}_b$. The detailed gradient estimation algorithm is shown in Alg.\ref{alg:grad-approx}. 
Note that if we let $\text{span}(W)=\mathbb{R}^m$, this step will be the same as in \cite{chen2019hopskipjumpattack}. However, we will sample from some representative subspace so that the gradient estimation is more efficient, and the corresponding theoretic justification is discussed in Section \ref{sec:dimred-theory}.
% In \cite{chen2019hopskipjumpattack} the authors sample uniformly in the unit ball in the entire input space $\mathbb{R}^m$. 

\begin{algorithm}
\caption{Gradient Approximation Based \name}
\label{alg:grad-approx}
\begin{algorithmic}[1]
\renewcommand{\algorithmicrequire}{\textbf{Input:}}
 \renewcommand{\algorithmicensure}{\textbf{Output:}}
 \REQUIRE a data point on the decision boundary ${\bf x} \in \mathbb{R}^m$, basis of the subspace $W \in \mathbb{R}^{m\times n}$, number of random sampling $B$, access to query the decision of victim model $\phi$.
 \ENSURE the approximated gradient $G$
 \STATE sample $B$ random Gaussian vectors of the lower dimension: $V_{rnd} \in \mathbb{R}^{B \times n}$.
 \STATE project the random vectors onto the gradient basis to get the perturbation vectors: $U_{rnd} = V_{rnd} \cdot W^\intercal$.
 \STATE get query points by adding perturbation vectors with the original point on the decision boundary: ${\bf x}_q[i] = {\bf x} + U_{rnd}[i]$.
%  \STATE query the victim model to get the binary decisions(1 for success, -1 otherwise): $D = g_{victim}(X_{query})$
 \STATE Monte Carlo approximation for the gradient: $G = \frac{1}{B}\sum_{i=1}^{B} \phi({\bf x}_q[i]) \cdot U_{rnd}[i]$
 \RETURN $G$
\end{algorithmic}
\end{algorithm}
\vspace{-0.5cm}

% \textbf{Difference on Gradient Estimation in Boundary-based Attack compared with Score-based Attack.} Note that both score-based and gradient-assisted boundary-based attacks do gradient estimation, but both the usage of estimated gradients and the capability of the attacker are different. 
% In score-based attacks, the attacker keep moving the image (which has the benign label) towards the estimated gradient direction until it is predicted as the adversarial label. In boundary-based attacks, the \advimage on the boundary is moved toward the gradient direction so that its adversarial prediction score is increased, and then project back to the boundary so that its distance towards the target-image is decreased.
% In score-based attacks, gradient estimation can be done at each data point with model's confidence scores, but the boundary-based attackers are only able to estimate gradients at the decision boundary using Monte Carlo methods. 
% \Huichen{todo:move}

\paragraph{Move along estimated gradient}
After we have estimated the gradient of adversarial prediction score $\nabla S$, we will move the $\xadv{t}$ towards the gradient direction:
\begin{align}
    \label{eqn:move-grad}
    \hat{\bf x}_{t+1} = \xadv{t} + \xi_t \cdot \frac{\widetilde{\nabla S}}{||\widetilde{\nabla S}||_2}
\end{align}
where $\xi_t$ is the step size at the $t$-th step. Hence, the prediction score of the adversarial class will be increased.

\vspace{-2mm}
\paragraph{Project to decision boundary}
% Now that the adversarial prediction score is larger than 0\bo{score or $\phi$? I think we should not talk about score any more in our method sinc it won't use it right? otherwise it would be confusing. we can introduce score for score based attack and be done. never mention it again in our method.}
Current $\hat{\bf x}_{t+1}$ is beyond the boundary, we can move the \advimage towards the target image so that it is projected back to the decision boundary:
\begin{align}
    \label{eqn:binary}
    \xadv{t+1} = \alpha_t \cdot {\bf x}_{tgt} + (1-\alpha_t) \cdot \hat{\bf x}_{t+1}
\end{align}
where the projection is achieved by a binary search over $\alpha_t$.

Note that we assume $\xadv{t}$ lies on the boundary while ${\bf x}_{src}$ does not lie on the boundary. Therefore, in the initialization step we need to first apply a project operation as in Eqn. \ref{eqn:binary} to get $\xadv{0}$.

In the following sections, we will introduce three exploration for the representative subspace optimization from spatial, frequency, and intrinsic component perspectives.

\subsection{Spatial Transformed Subspace (\name-S)}
\label{sec:name-S}
First we start with the spatial transformed query reduction approach.
The intuition comes from the observation that the gradient of input image has a property of local similarity\cite{ilyas2018prior}. Therefore, a large proportion of the gradients lies on the low-dimensional subspace spanned by the bilinear interpolation operation\cite{spath1993two}.
% The intuition comes from image downsampling where a group of image pixels in a rectangle space is merged into one `hyper pixel' in the new image, and we use the inverse process here.
In order to sample random perturbations for an image, we first sample a lower-dimensional random perturbation $Q$ of shape $\lfloor \frac Nr \rfloor \times \lfloor \frac Nr \rfloor$, where $r$ is the hyperparameter of dimension reduction factor. Then we use bilinear-interpolation to map it back the original image space, $X = \text{Bil\_Interp}(Q)$.

The basis of this spatial transformed subspace is the images transformed from unit perturbations in the lower space:
\begin{align*}
    w^{(i,j)} = \text{Bil\_Interp}(e^{(i,j)}),\quad 0\leq i,j \leq \lfloor N/r \rfloor
\end{align*}
where $e^{(i,j)}$ represents the unit vector that has 1 on the $(i,j)$-th entry and 0 elsewhere.
% In order to sample random perturbations for an image, we sample a smaller vector $v_h$ where each value is the perturbation for a group of pixels in the image. Let the group size be $r_g > 0$, $v_h \sim \mathbb{R}^{\frac{m}{r_g}}$. Then we map $v_h$ into $\mathbb{R}^{m}$ by copying each value for $r_g$ times to fill the pixels in the group.
% \bo{remember to cite the algorithm if you put in the paper}

\subsection{Low Frequency Subspace (\name-F)}
\label{sec:name-F}
In general the low frequency subspace of an image contains the most of the critical information, including the gradient information\cite{guo2018low}; while the high frequency signals contain more noise than useful content.
Hence, we would like to sample our perturbations from the low frequency subspace via Discrete Cosine Transformation(DCT)\cite{ahmed1974discrete}. Formally speaking, define the basis function of DCT as:
\begin{align}
    \phi(i,j) = \cos \bigg(\frac{(i+\frac12)j}{N} \pi \bigg)
\end{align}
The inverse DCT transformation is a mapping from the frequency domain to the image domain $X=\text{IDCT}(Q)$:
\begin{align}
    X_{i_1,i_2}=\sum_{j_1=0}^{N-1}\sum_{j2=0}^{N-1} N_{j_1} N_{j_2} Q_{j_1,j_2} \phi(i_1,j_1) \phi(i_2,j_2)
\end{align}
where $N_j=\sqrt{1/N}$ if $j=0$ and otherwise $N_j=\sqrt{2/N}$.

We will use the lower $\lfloor N/r \rfloor$ part of the frequency domain as the subspace, i.e.
\begin{align}
    w^{(i,j)} = \text{IDCT}(e^{(i,j)}),\quad 0\leq i,j \leq \lfloor N/r \rfloor
\end{align}
where hyperparameter $r$ is the dimension reduction factor.
% Here $r$ denotes the hyperpameter of dimension reduction factor. For example, if we use $r=4$, we get a subspace whose dimension is $1/16$ of that from the original space.

% The basis of the frequency domain is the images transformed from unit vectors in the frequency space:
% \begin{align}
%     w^{(i,j)} = \text{IDCT}(e^{(i,j)})
% \end{align}
% where $e^{(i,j)}$ represents the unit vector that has 1 on the $(i,j)$-th entry and 0 elsewhere. We use the first $\lfloor N/r \rfloor$ part as the basis of the low frequency subspace:
% \begin{align}
%     \{w^{(i,j)}\},\quad 0\leq i,j \leq \lfloor N/r \rfloor
% \end{align}

% Discrete Cosine Transformation (DCT) can transform an image to frequency domain and is used for image compression by only keeping the low-frequency components~\cite{wallace1992jpeg}. Here we first sample a low-dimension random vector $v_f \in \mathbb{R}^{r_f}$ as the perturbations for low-frequency components. To map this vector to the original space, we append 0's to it as the perturbations for higher-frequency components, then do an inverse DCT to get back to image domain.

\subsection{Intrinsic Component Subspace (\name-I)}
\label{sec:name-I}
% \Xiaojun{gradient space, transferbility, randomized, disk}
\begin{figure}
    \centering
    \includegraphics[width=\linewidth]{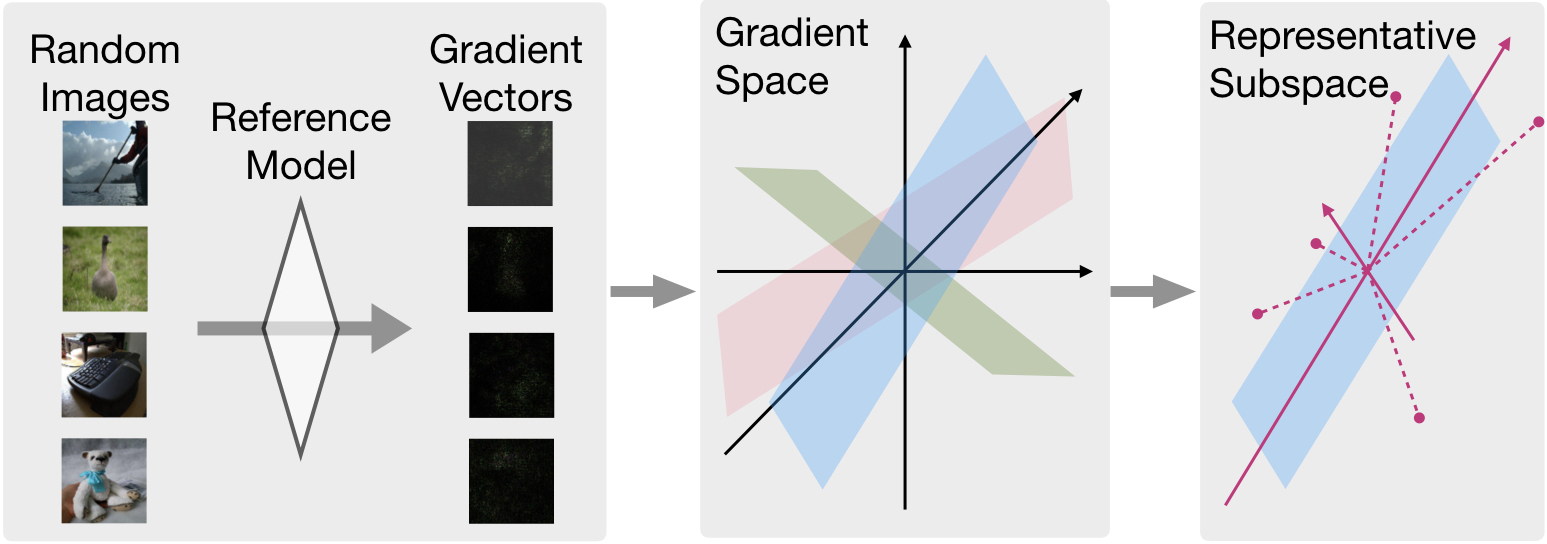}
    \caption{Generate representative subspace from the original high-dimensional gradient space.}
    \label{fig:bases_generation}
    \vspace{-0.5cm}
\end{figure}

Principal Component Analysis (PCA)\cite{wold1987principal} is a standard way to perform dimension reduction in order to search for the intrinsic components of the given instances. Given a set of data points in high dimensional space, PCA aims to find a lower dimensional subspace so that the projection of the data points onto the subspace is maximized. 

Therefore, it is possible to leverage PCA to optimize the subspace for model gradient matrix. However, in order to perform PCA we will need a set of data points. In our case that should be a set of gradients of $S({\bf x})$ w.r.t. different $\bf x$. This is not accessible under black-box setting. Hence, we turn to a set of `reference models' to whose gradient we have access. As shown in Figure \ref{fig:bases_generation}, we will use a reference model to calculate a set of image gradients ${\bf g}_1, {\bf g}_2, \ldots, {\bf g}_K \in \mathbb{R}^m$ Then we perform a PCA to extract its top-$n$ principal components - ${\bf w}_1, \ldots, {\bf w}_n \in \mathbb{R}^m$. These $w$'s are the basis of the Intrinsic Component Subspace.
Note that different from transferability, we do not restrict the reference models to be trained by the same training data with the original model, since we only need to search for the intrinsic components of the give dataset which is relatively stable regarding diverse models.
% By fitting an optimal subspace over gradients of different model, we can find the subspace that captures the the gradient space on the task for all the models. 
% The detailed algorithm is shown in Alg. \ref{alg:basis-gen}.

In practice, the calculation of PCA may be challenging in terms of time and memory efficiency based on large high-dimensional dataset (the data dimension on ImageNet is over 150k and we need a larger number of data points, all of which are dense). Therefore, we leverage the randomized PCA algorithms\cite{halko2011finding} which accelerates the speed of PCA while achieving comparable performance.
% In addition, we can further improve time efficiency by noticing that we can calculate the subspace of the top-$K$ components of PCA without calculating the exact value of the top-$K$ components. The detailed algorithm is as follows\Xiaojun{todo}.

An additional challenge is that the matrix $X$ may be too large to be stored in memory. Therefore, we store them by different rows since each row (i.e. gradient of one image) is calculated independently with the others. The multiplication of $X$ and other matrices in memory are then implemented accordingly.

\section{Theoretic Analysis on \name}
% \subsection{Dimension Reduction Theorem}
% \section{Dimension Reduction Helps Gradient Estimation}
\label{sec:dimred-theory}

% We propose to apply dimension reduction techniques to improve the gradient estimation quality. The intuition is straightforward: if the dimension of the space from which we sample the $u$'s is smaller, then the estimation quality will be better. Hence, we will choose a subspace in $\mathbb{R}^m$ from which we sample the random perturbations.
% We theoretically show that the gradient estimation quality will be better if we can sample from an informative subspace rather than from the entire space.
% In this section, w
We theoretically analyze how dimension reduction helps with the gradient estimation in \name. We show that the gradient estimation bound is tighter by sampling from a representative subspace rather than the original space.

We consider the gradient estimation as in Eqn. \ref{eq:MC_gradient_estimation} and let $\rho = \frac{||\text{proj}_{\text{span}(W)}(\nabla S)||_2}{||\nabla S||_2}$ denote the proportion of $\nabla S$ that lies on the chosen subspace $\text{span}(W)$. Then we have the following theorem on the expectation of the cosine similarity between $\nabla S$ and estimated $\widetilde{\nabla S}$:

\begin{theorem}
\label{tho:dimred}
Suppose 1) $S({\bf x})$ has $L$-Lipschitz gradients in a neighborhood of $\bf x$, 2) the sampled ${\bf v}_1, \ldots, {\bf v}_B$ are orthogonal to each other, and 3) $W^\intercal W = I$, then the expected cosine simliarity between $\widetilde{\nabla S}$ and $\nabla S$ can be bounded by:
% For a boundary point $x$, suppose that 1) $S(x)$ has $L$-Lipschitz gradients in a neighborhood of $x$, 2) the sampled ${\bf v}_1, \ldots, {\bf v}_B$ are orthogonal to each other, 3) $W^\intercal W = I$. Let constant $\rho = \frac{||\text{proj}_{\text{span}(W)}(\nabla S)||_2}{||\nabla S||_2}$ denote the proportion of $\nabla S$ that lies on $\text{span}(W)$. Then the expected cosine simliarity between $\widetilde{\nabla_W y}$ and $\nabla y$ can be bounded by:
\begin{align}
    &\bigg( 2\bigg(1-(\frac{L\delta}{2||\nabla S||_2})^2\bigg)^{\frac{n-1}{2}} - 1 \bigg)c_n\rho\sqrt{\frac B n} \\
    \leq & \mathbb{E}\big[\cos (\widetilde{\nabla S}, \nabla S) \big]\\
    \leq & c_n\rho\sqrt{\frac B n}
\end{align}
where $c_n$ is a coefficient related with the subspace dimension $n$ and can be bounded by $c_n \in (2/\pi, 1)$. In particular:
\begin{equation}
    \label{eqn:grad-est-qual-dr}
    \lim_{\delta\rightarrow 0}\mathbb{E}\big[\cos (\widetilde{\nabla_W S}, \nabla S) \big] = c_n\rho 
    \sqrt{\frac B n}.
\end{equation}
\end{theorem}
The theorem proof is in Appendix \ref{sec:tho-proof}. 
If we sample from the entire space (i.e. $\text{span}(W) = \mathbb{R}^m$), the expected cosine similarity is $c_m\sqrt{\frac{B}{m}}$.
If we let $m=3\times224\times224$ and $B=100$, the similarity is only around 0.02.
% Hence, we claim that gradient estimation by sampling from the entire space is not an efficient estimation approach.

On the other hand, if the subspace basis $w$'s are randomly chosen, then $\rho \approx \sqrt{\frac{n}{m}}$ and the estimation quality is low. With larger $\rho$, the estimation quality will be better than sampling from the entire space.
Therefore, we further explore three approaches to optimize the representative subspace that contains a larger portion of the gradient as discussed in Section~\ref{sec:subspaces}.
For example, in the experiments we see that when $n=m/16$, we can reach $\rho=0.5$ and the expected cosine similarity increase to around 0.06.
This improves the gradient estimation quality which leads to more efficient attacks.
% with a carefully chosen subspace with more information, we can have a larger $\rho$ to achieve better estimation quality of the gradient. Thus, the key is to choose a subspace that maximizes $\rho$.

\section{Experiments}
In this section, we introduce our experimental setup and quantitative results of the proposed methods \name-S, \name-F, and \name-I, compared with the HSJA attack\cite{chen2019hopskipjumpattack}, which is the-state-of-the-art boundary-based blackbox attack.
Here we focus on the strongest baseline HSJA, which outperforms all of other Boundary Attack~\cite{brendel2017decision}, Limited Attack~\cite{ilyas2018black} and Opt Attack~\cite{cheng2018query} by a substantial margin.
We also show two sets of qualitative results for attacking two real-world APIs with the proposed methods.
\subsection{Datasets and Experimental Setup}
\paragraph{Datasets}
We evaluate the attacks on two offline models on ImageNet\cite{deng2009imagenet} and CelebA\cite{liu2018large} and two online face recognition APIs Face++\cite{facepp-compare-api} and Azure\cite{azure-detect-api}. We use a pretrained ResNet-18 model as the target model for ImageNet and fine-tune a pretrained ResNet-18 model to classify among 100 people in CelebA. We randomly select 50 pairs from the ImageNet/CelebA validation set that are correctly classified by the model as the source and target images. 
\paragraph{Attack Setup}
Following the standard setting in \cite{chen2019hopskipjumpattack}, we use $\xi_t=||\xadv{t-1}-{\bf x}_{tgt}||_2/\sqrt{t}$ as the size in each step towards the gradient. We use $\delta_t=\frac1m ||\xadv{t-1}-{\bf x}_{tgt}||_2$ as the perturbation size and $B=100$ queries in the Monte Carlo algorithm to estimate the gradient, where $m=3\times224\times224$ is the input dimension in each Monte Carlo step.

We provide two \textbf{evaluation metrics} to evaluate the attack performance. The first is the average Mean Square Error (MSE) curve between the target image and the adversarial example in each step, indicating the magnitude of perturbation. The smaller the perturbation is, the more similar the adversarial example is with the \targetimage, thus providing better attack quality.
The second is the attack success rate based on a limited number of queries, where the `success' is defined as reaching certain specific MSE threshold. The less queries we need in order to reach a certain perturbation threshold, the more efficient the attack method is.

% In CIFAR, we use a factor of 2 in resize and DCT, which gives a 768 dimensional subspace. We extract the top 768 major components in PCA.
As for the dimension-reduced subspace, we use the dimension reduction factor $r=4$ 
% \bo{define this thing? give a notation maybe? check the statements to be rigrous!!}\Xiaojun{defined this in section 4.1, 4.2}
in \emph{spatial transformed} and \emph{low frequency} subspace, which gives a 9408 dimensional subspace.
In order to generate the \emph{Intrinsic Component Subspace}, we first generate a set of image gradient vectors on the space. We average over the gradient of input w.r.t. five different pretrained substitute models - ResNet-50\cite{he2016deep}, DenseNet-121\cite{huang2017densely}, VGG16\cite{simonyan2014very}, WideResNet\cite{zagoruyko2016wide} and GoogleNet\cite{szegedy2015going}. We use part of the ImageNet validation set (280000 images) to generate the gradient vectors. Finally we adopt the scalable approximate PCA algorithm\cite{halko2011finding} to extract the top 9408 major components as the intrinsic component subspace.

% \bo{in this above section, please separately discuss the setting for the three methods. this is an important part.}

% We use the approximate PCA algorithm \Xiaojun{cite} to extract the top 9408 major components to improve efficiency.
% In order to generate the gradient vectors for PCA, we will average over the gradient of input w.r.t. five different pretrained substitute models - ResNet-50, DenseNet-121, VGG16, WideResNet and GoogleNet\Xiaojun{cite}. We use part of the ImageNet validation set (280000 images) to generate the gradient vectors.
% Other attack parameter setting is the same as that in \cite{chen2019hopskipjumpattack}.

\subsection{Commercial Online APIs}
Various companies provide commercial APIs (Application Programming Interfaces) of trained models for different tasks such as face recognition. Developers of downstream tasks can pay for the services and integrate the APIs into their applications. Note that although typical platform APIs provide the developers the confidence score of classes associated with their final predictions, the end-user using the final application would not have access to the scores in most cases. For example, some of Face++'s partners use the face recognition techniques for log-in authentication in mobile phones~\cite{facepp-partner}, where the user only knows the final decision (whether they pass the verification or not).

We choose two representative platforms for our real-world experiments based on only the final prediction. The first is Face++ from MEGVII\cite{facepp-compare-api}, and the second is Microsoft Azure\cite{azure-detect-api}. Face++ offers a `compare' API~\cite{facepp-compare-api} with which we can send an HTTPS request with two images in the form of byte strings, and get a prediction confidence of whether the two images contain the same person. In all the experiments we consider a confidence greater than 50\% meaning the two images are tagged as the same person. Azure has a slightly more complicated interface. To compare two images, each image first needs to pass a `detect' API call~\cite{azure-detect-api} to get a list of detected faces with their landmarks, features, and attributes. Then the features of both images are fed into a `verify' function~\cite{azure-verify-api} to get a final decision of whether they belong to the same person or not. The confidence is also given, but we do not need it for our experiments since we only leverage the binary prediction for practical purpose.

\begin{figure}
\centering
\begin{subfigure}[t]{.28\linewidth}
  \centering
  \includegraphics[width=\linewidth]{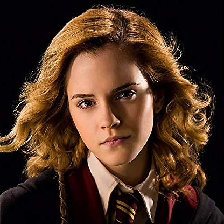}
  \caption{Person 1}
  \label{fig:tgt_img}
\end{subfigure}
\begin{subfigure}[t]{.28\linewidth}
  \centering
  \includegraphics[width=\linewidth]{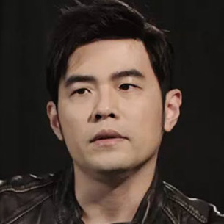}
  \caption{Person 2}
  \label{fig:src_img_facepp}
\end{subfigure}
\begin{subfigure}[t]{.28\linewidth}
  \centering
  \includegraphics[width=\linewidth]{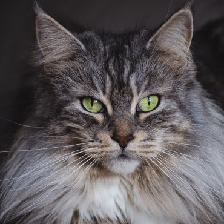}
  \caption{No-face}
  \label{fig:src_img_azure}
\end{subfigure}
\caption{
The source and target images for online API experiments. All images are resized to $3\times 224\times 224$. 
Image~\ref{fig:tgt_img} is the \targetimage for both APIs. Image~\ref{fig:src_img_facepp} is the \sourceimage for attacking Face++ `compare' API, and \ref{fig:src_img_azure} the \sourceimage for Azure `detect' API. }
\label{fig:src_tgt_imgs}
% \vspace{-0.5cm}
\end{figure}

In the experiments, we use the examples in Figure~\ref{fig:src_tgt_imgs} as \sourceimage and \targetimage. More specifically, we use a man-woman face as the source-target pair for the `compare' API Face++, and we use a cat-woman face as the pair for the `detect' API Azure face detection.

\begin{figure*}[htpb]
\begin{subfigure}[t]{0.2465\linewidth}
    \centering
    \includegraphics[width=\textwidth]{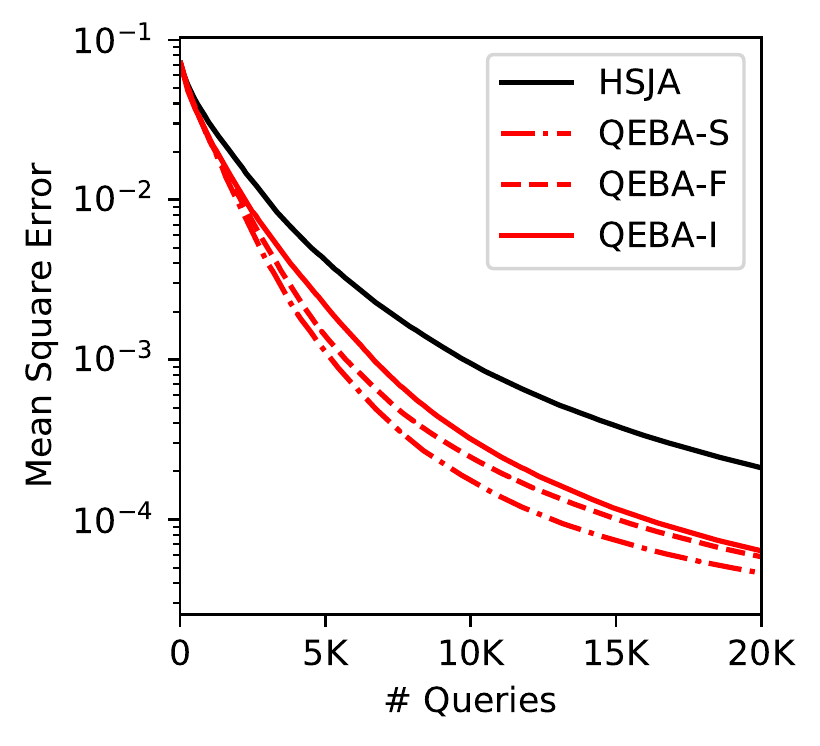}
    \caption{The MSE vs. query number on ImageNet.}
    \label{fig:result-imagenet}
\end{subfigure}
\hspace{1mm}
\begin{subfigure}[t]{0.231\linewidth}
    \centering
    \includegraphics[width=\textwidth]{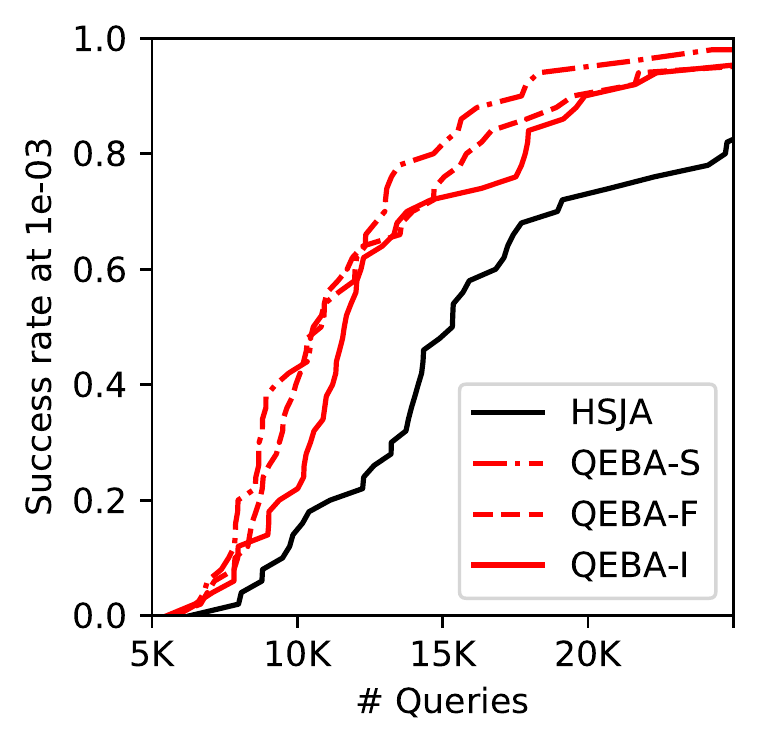}
    \caption{The attack success rate with threshold $10^{-3}$ on ImageNet.}
    \label{fig:result-imagenet-sr}
\end{subfigure}
\hspace{1mm}
\begin{subfigure}[t]{0.2465\linewidth}
    \centering
    \includegraphics[width=\textwidth]{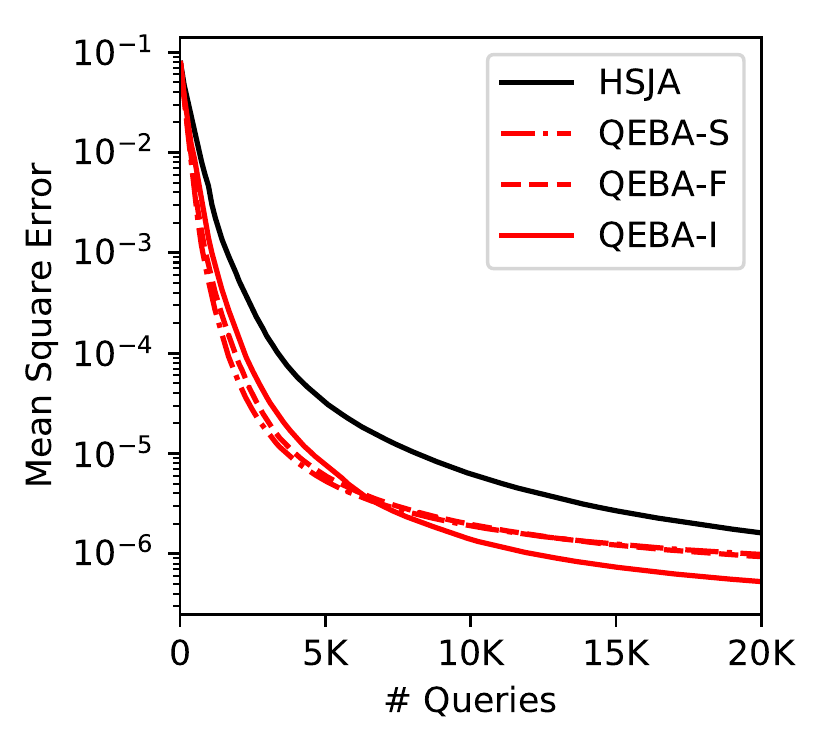}
    \caption{The MSE vs. query number on CelebA.}
    \label{fig:result-celeba}
\end{subfigure}
\hspace{1mm}
\begin{subfigure}[t]{0.235\linewidth}
    \centering
    \includegraphics[width=\textwidth]{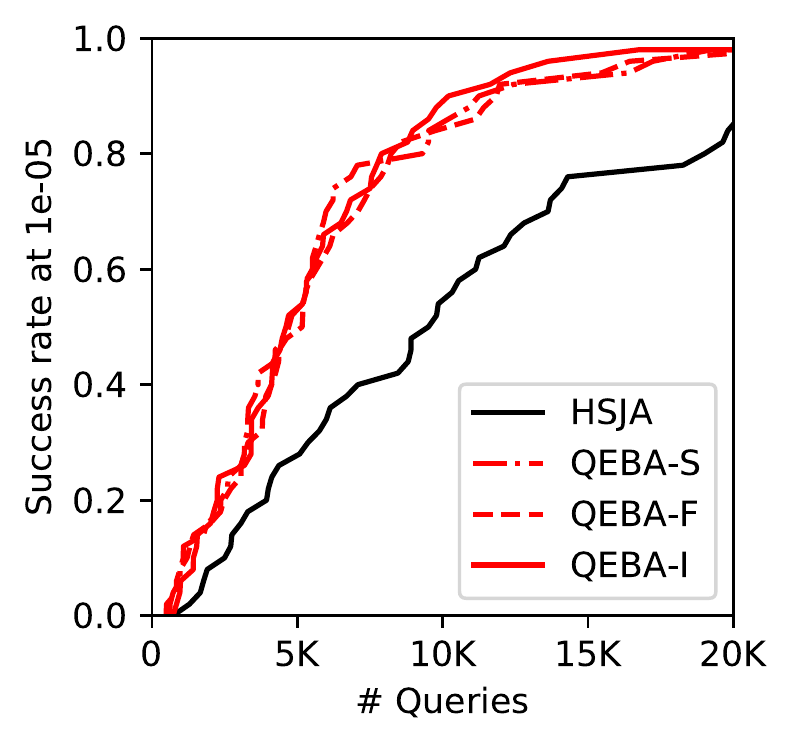}
    \caption{The attack success rate with threshold $10^{-5}$ on CelebA.}
    \label{fig:result-celeba-sr}
\end{subfigure}
\caption{The attack results on ImageNet and CelebA datasets.}
\label{fig:results}
\end{figure*}

% \subsection{Optimizations for Experiments}
\paragraph{Discretization Optimization for Attacking APIs}
The attack against online APIs suffers from the problem of `discretization'. That is, in the attack process we assume the pixel values to be continuous in $[0,1]$, but we need to round it into 8-bit floating point in the uploaded RGB images when querying the online APIs.
% The pixel values of each channel in RGB images are 8-bit floating point numbers and the pixel array of an image takes the value of $\mathbb{D}^m \in \{0,\frac{1}{255},\ldots,\frac{254}{255},1\}^m$. There are at most $256^3$ colors possible.
% A typical computer, on the other hand, has 32-bit or 64-bit system. So the querying samples generated by boundary-based attacks with small perturbation around the decision boundary for gradient estimation are likely to contain values that cannot be represented with only 8 bits. We refer to these 32-bit or 64-bit pixel values as `continuous' and the 8-bit images as `discrete' in our discussion for simplicity.
% This is not a problem for offline experiments, since a local machine learning model for image recognition does not set constraint on the input range and it can work as normal on a `continuous image'.
% However, the difference between value ranges incurs a problem for attacking online APIs. The commercial platforms require the uploaded images to have valid pixel values, so the perturbed \queryimages have to be rounded.
This would cause error in the Monte Carlo gradient estimation format in Equation~\ref{eq:MC_gradient_estimation} since the real perturbation between the last \boundaryimage and the new \queryimage after rounding is different from the weighted perturbation vector $\delta {\bf u}_b$. 

In order to mitigate this problem, we perform discretization locally. Let $P_{rd}$ be a projection from a continuous image $\bf x_c$ to a discrete image $\bf x_d = P_{rd}(\bf x_c)$. Let $\delta {\bf u'}_b = P_{rd}(\bf x+\delta {\bf u}_b) - x$, the new gradient estimation format becomes:
\begin{align}
    \widetilde{\nabla f} &= \frac1B \sum_{i=1}^B \phi(P_{rd}({\bf x}+\delta {\bf u}_b)) {\bf u'}_b.
\end{align}

% \bo{this section should be moved to sub of commercial APIs maybe}\Huichen{Ok. Done.}

% One obstacle in applying \Xiaojun{our approach} to attack real-world machine learning system is the problem of \emph{discretization}. That is to say, during our attack process the adversarial examples are considered in the continuous space $\mathbb{R}^m$ (or $[0,1]^m$ if we bound the pixel values), but in reality image pixel values are rounded to an 8-bit float number $\mathbb{D}^m = \{0,\frac{1}{255},\ldots,\frac{254}{255},1\}^m$. 
% Boundary-based attack relies on querying with small perturbation around the decision boundary to estimate the gradient, so it is sensitive to the problem of discretization.

% In order to mitigate the effect of rounding, we propose a pipeline to discretize the subspace from which we are sampling. In particular, suppose the \advimage at the current step is $\xadv{t}$ and we originally want to sample from the subspace ${\bf u} \sim \text{span}(W)$. In a discretized setting, we will discretize $\text{span}(W)$ into $Discr(\text{span}(W)\big|\xadv{t})$ such that ${\bf u} \sim Discr(\text{span}(W)\big|\xadv{t})$ will satisfy:
% \begin{align}
%     \label{eqn:discr}
%     \xadv{t}+\delta{\bf u} \in \mathbb{D}^m
% \end{align}
% The intuition of this approach is that Eqn.\ref{eqn:discr} ensures that our query to the model ($\phi(\xadv{t}+\delta{\bf u}_b$) will not be changed in the rounding process, so we can get a more accurate gradient estimation compared with that when we do not use the discretized subspace.

\subsection{Experimental Results on Offline Models}

\begin{table*}[ht]
  \centering
  \caption{Attack success rate using different number of queries and different MSE thresholds.}
%   \vspace{-0.2cm}
    \renewcommand\tabcolsep{5pt} % adjust the space between each column 
    \begin{threeparttable}
    \begin{tabular}{l|l|cccc|cccc|cccc}
    \toprule
    &  & \multicolumn{4}{c|}{\# Queries = 5000} & \multicolumn{4}{c|}{\# Queries = 10000} & \multicolumn{4}{c}{\# Queries = 20000} \\
    \cmidrule{2-6} \cmidrule{7-10} \cmidrule{11-14} 
    & \makecell{MSE\\ threshold} & \makecell{HJSA} & \makecell{\\-S} & \makecell{\name\\-F} & \makecell{\\-I} & \makecell{HJSA} & \makecell{\\-S} & \makecell{\name\\-F} & \makecell{\\-I} & \makecell{HJSA} & \makecell{\\-S} & \makecell{\name\\-F} & \makecell{\\-I} \\ 
    % \midrule
    \cmidrule{1-2}\cmidrule{3-6} \cmidrule{7-10} \cmidrule{11-14} 
%     \multirow{3}[0]{*}{Cifar10} 
%     & 0.01 & 1.00 & 1.00 & 0.94 & 0.76 & 0.16 & 0.02 & 0.96 & 0.90 & 0.76  \\
% 	& 0.001 & 1.00 & 1.00 & 0.60 & 0.86 & 0.40 & 0.08 & 1.00 & 1.00 & 0.96\\
% 	& 0.0001 & 1.00 & 1.00 & 0.68 & 0.86 & 0.42 & 0.06 & 1.00 & 1.00 & 0.96\\
% 	\midrule
    \multirow{3}[0]{*}{ImageNet} 
    & 0.01 & 0.76 & \bf 0.86 & \bf 0.86 & \bf 0.86 & 0.98 & \bf 1.00 & 0.96 & 0.98 & \bf 1.00 & \bf 1.00 & \bf 1.00 & \bf 1.00 \\
	& 0.001 & 0.16 & 0.40 & \bf 0.42 & 0.36 & 0.50 & 0.74 & \bf 0.76 & 0.74 & 0.84 & \bf 0.98 & 0.96 & \bf 0.98 \\
	& 0.0001 & 0.02 & \bf 0.08 & 0.06 & 0.04 & 0.06 & \bf 0.32 & 0.30 & 0.20 & 0.28 & \bf 0.70 & 0.66 & 0.68\\
	\midrule
    \multirow{3}[0]{*}{CelebA} 
    & 0.01 & 0.96 & \bf 1.00 & \bf 1.00 & 0.96 & \bf 1.00 & \bf 1.00 & \bf 1.00 & \bf 1.00 & \bf 1.00 & \bf 1.00 & \bf 1.00 & \bf 1.00 \\
	& 0.001 & 0.90 & \bf 1.00 & \bf 1.00 & 0.94 & 0.98 & \bf 1.00 & \bf 1.00 & \bf 1.00 & \bf 1.00 & \bf 1.00 & \bf 1.00 & \bf 1.00 \\
	& 0.0001 & 0.76 & \bf 0.96 & \bf 0.96 & 0.90 & 0.90 & \bf 1.00 & \bf 1.00 & \bf 1.00 & \bf 1.00 & \bf 1.00 & \bf 1.00 & \bf 1.00 \\
    \bottomrule
    \end{tabular}%
    \end{threeparttable}
  \label{tab:results}%
\end{table*}

\begin{figure}
    \centering
    \includegraphics[width=0.95\linewidth]{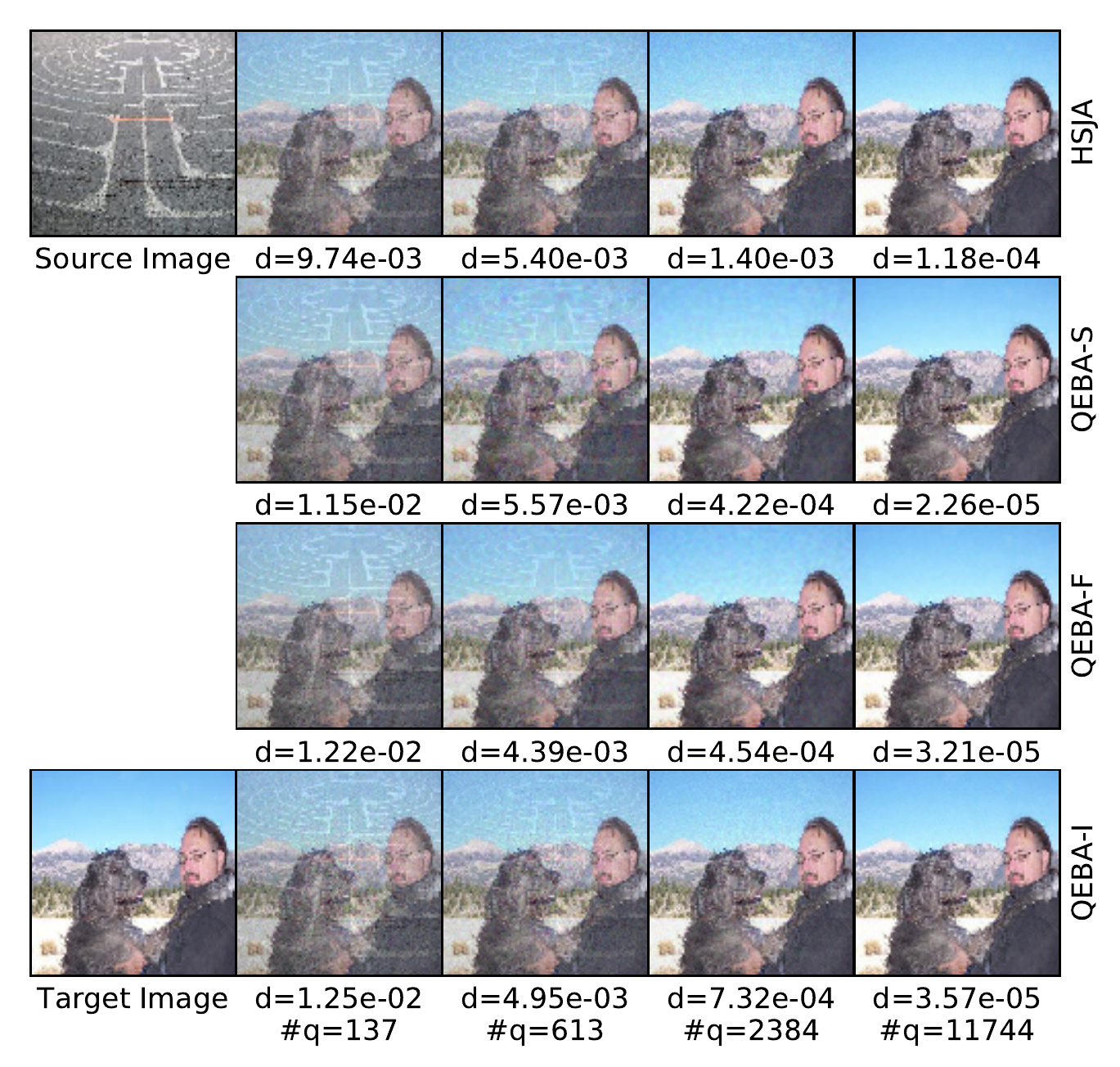}
    \caption{An example of attacking ImageNet trained model based on different subspaces.}
    \label{fig:result-process}
    \vspace{-0.5cm}
\end{figure}
% \Xiaojun{Talk about cifar}CIFAR: Figure\ref{fig:result-cifar}; maybe remove it?

To evaluate the effectiveness of the proposed methods, we first show the average MSE during the attack process of ImageNet and CelebA using different number of queries in Figure~\ref{fig:result-imagenet} and Figure \ref{fig:result-celeba} respectively. 
We can see that all the three proposed query efficient methods outperform HSJA significantly.
We also show the attack success rate given different number of queries in Table \ref{tab:results} using different MSE requirement as the threshold. In addition, we provide the attack success rate curve in Figure \ref{fig:result-imagenet-sr} and \ref{fig:result-celeba-sr} using $10^{-3}$ as the threshold for ImageNet and $10^{-5}$ for CelebA to illustrate convergence trend for the proposed \name-S, \name-F, and \name-I, comparing with the baseline HJSA.

We observe that sampling in the optimized subspaces results in a better performance than sampling from the original space. The spatial transforamed subspace and low-frequency subspace show a similar behaviour since both of them rely on the local continuity. The intrinsic component subspace does not perform better than the other two approaches, and the potential reason is that we are only using 280000 cases to find intrinsic components on the 150528-dimensional space. Therefore, the extracted components may not be optimal. We also observe that the face recognition model is much easier to attack than the ImageNet model, since the face recognition model has fewer classes (100) rather than 1000 as of ImageNet. 

A qualitative example process of attacking the ImageNet model using different subspaces is shown in Figure \ref{fig:result-process}. In this example, the MSE (shown as $d$ in the figures) reaches below $1\times10^{-3}$ using around 2K queries when samlping from the subspaces, and it is already hard to tell the adversarial perturbations in the examples. When we further tune the \advimage using 10K queries, it reaches lower MSE.

% The attack result on ImageNet and CelebA is shown in Figure \ref{fig:results} and Table \ref{tab:results}. We see that sampling in all the subspaces results in a better performance than sampling from the entire one. The hyper-pixel subspace and frequency subspace show a similar behaviour since both of them rely on the local continuity. The intrinsic component subspace does not perform better than the other two simpler approaches. We owe it to the reason that we are only using 280000 cases to find intrinsic components on the 150528-dimensional space. Therefore, the used components may not be optimal. An example process of attacking the ImageNet model using different subspaces is shown in Figure \ref{fig:result-process}.

\subsection{Results of Attacking Online APIs}
\begin{figure}[t]
    \centering
    \includegraphics[width=0.91\linewidth]{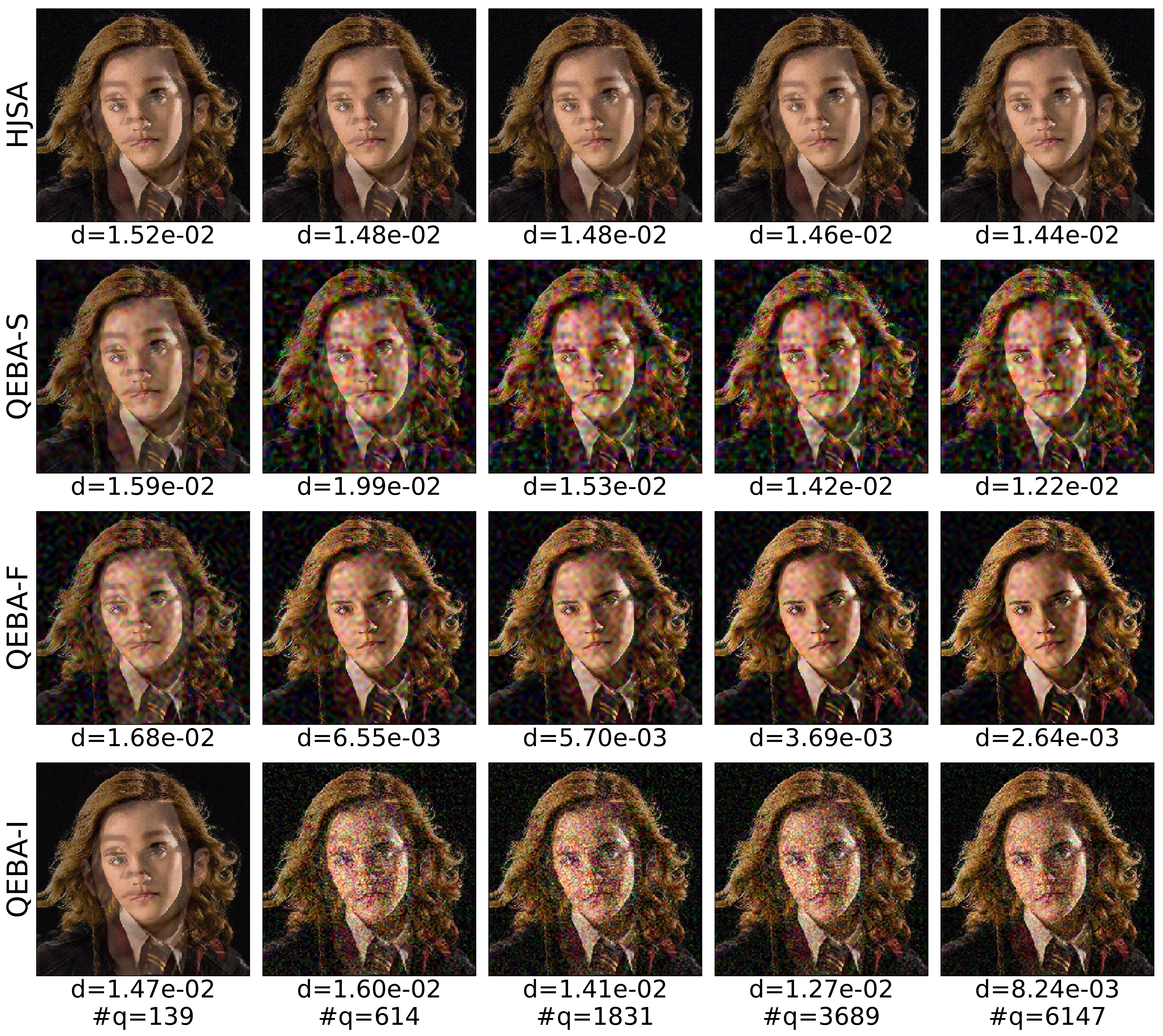}
    % \vspace{-0.1cm}
    \caption{Comparison of attacks on Face++ `compare' API. Goal: obtain an image that is tagged as `same person' with the \sourceimage person 2 (Figure~\ref{fig:src_img_facepp}) by the API when humans can clearly see person 1 here.}
    \label{fig:facepp_atk}
    \vspace{-0.3cm}
\end{figure}

\begin{figure}[t]
    \centering
    \includegraphics[width=0.91\linewidth]{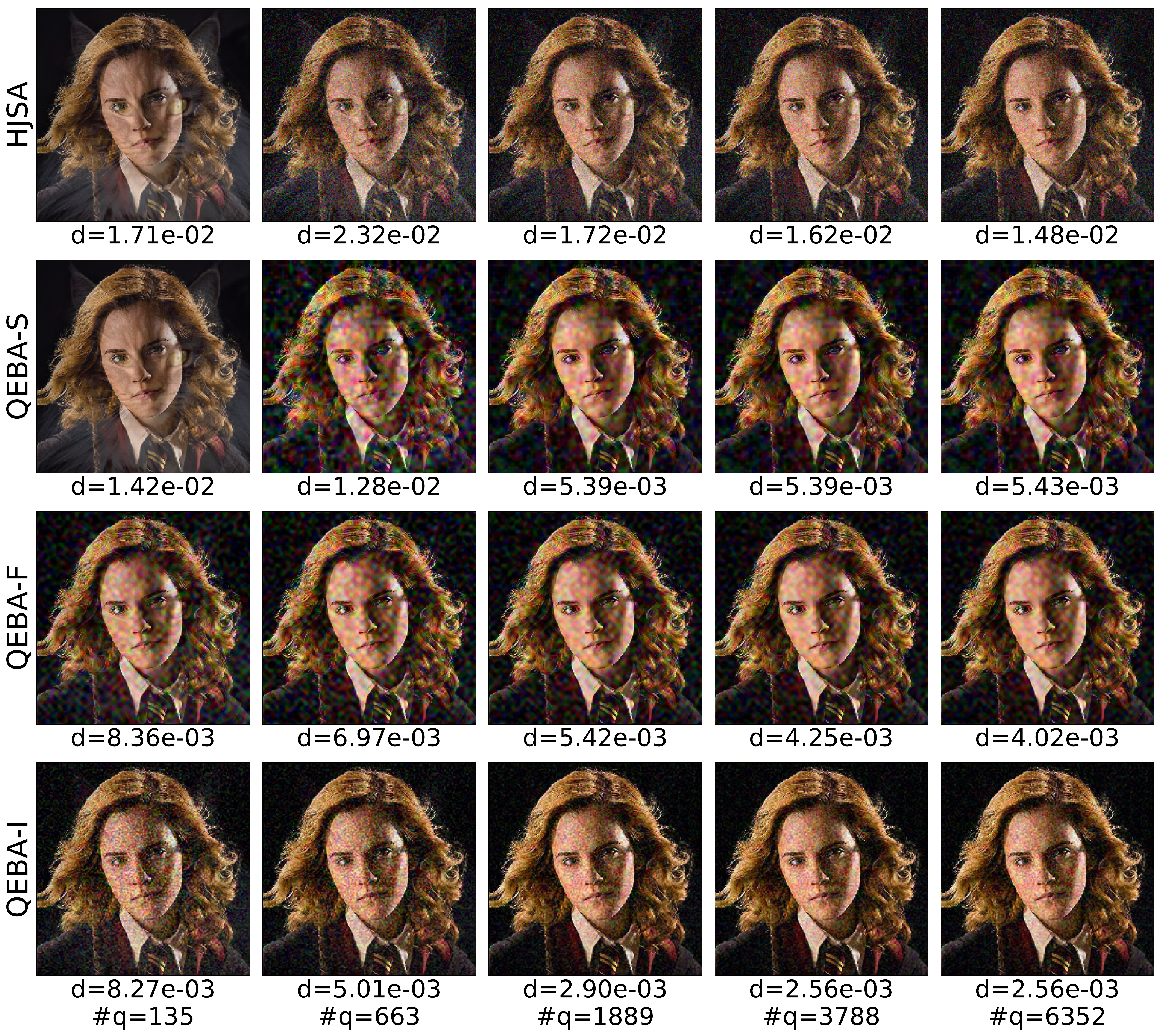}
    % \vspace{-0.1cm}
    \caption{Comparison of attacks on Azure `detect' API. Goal: get an image that is tagged as `no face' by the API when humans can clearly see a face there. The \sourceimage is a cat as shown in Figure~\ref{fig:src_img_azure}.}
    \label{fig:azure_atk}
    \vspace{-0.3cm}
\end{figure}

The results of attacking online APIs Face++ and Azure are shown in Figure~\ref{fig:facepp_atk} and Figure~\ref{fig:azure_atk} respectively.
% against Face++ and Azure face detection API. 
The labels on the y-axis indicate the methods. Each column represents successful attack instances with increasing number of API calls. \Huichen{I unify the two (.) and move front}
% (the perturbation magnitude $d$ is also shown). 
As is the nature of boundary-based attack, all images are able to produce successful attack. The difference lies in the quality of attack instances. 
% (perturbation magnitude $d$). 

For attacks on Face++ `compare' API, the \sourceimage is a man and the \targetimage is a woman as shown in Figure~\ref{fig:src_tgt_imgs}. Notice the man's eyes appear in a higher position in the \sourceimage than the woman in the \targetimage because of the pose. 
% The first column in Figure~\ref{fig:facepp_atk} show the results with a few queries, and we can see all methods produce images with four eyes in them at the beginning.
All the instances on the first row in Figure~\ref{fig:facepp_atk} based on HJSA attacks contain two pairs of eyes. The MSE scores ($d$ in the figures) also confirm that the distance between the attack instance and the \targetimage does not go down much even with more than 6000 queries. On the other hand, our proposed methods \name- can optimize attack instances with smaller magnitude of perturbation more efficiently. The perturbations are also smoother.

The attack results on Azure `detect' API show similar observations. The \sourceimage is a cat and the \targetimage is the same woman. Sampling from the original high-dimensional space (HJSA) gives us attack instances that presents two cat ear shapes at the back of the human face as shown in the first row in Figure~\ref{fig:azure_atk}. With the proposed query efficient attacks, the perturbations are smoother. The distance metric ($d$) also demonstrates the superiority of the proposed methods.

% this is for the arxiv version
% the cvpr version does not have the two \vspace commands and the width is 0.95 for both images

\section{Related Work}
% adv
% blackbox
% query efficient score-based blackbox
% a. more info; b. not guarantee success c. #query 

\paragraph{Boundary-based Attack}
% A line of work that is most related to ours is using decision-based attacks.
Boundary Attack~\cite{brendel2017decision} is one of the first work that uses final decisions of a classifier to perform blackbox attacks. The attack process starts from the \sourceimage, which is classified as the adversarial \maliciousclass. Then it employs a reject sampling mechanism to find a \boundaryimage that still belongs to the \maliciousclass by performing random walk along the boundary. The goal is to minimize the distance between the \boundaryimage and the \targetimage. However, as the steps taken are randomly sampled, the convergence of this method is slow and the query number is large.

Several techniques have been proposed to improve the performance of Boundary Attack. \cite{brunner2019guessing,srinivasan2019black,guo2018low} propose to choose the random perturbation in each step more wisely instead of Gaussian perturbation, using Perlin noise, alpha distribution and DCT respectively. \cite{ilyas2018black,khalid2019red,liu2019geometry,chen2019hopskipjumpattack} propose a similar idea - approximating the gradient around the boundary using Monte Carlo algorithm. 
% In our work, we mainly follow the pipeline in \cite{chen2019hopskipjumpattack} because it establishes a good theoretic analysis for the model.

There are two other blackbox attacks which are not based on the boundary. \cite{cheng2018query} proposes to transform the boundary-based output into a continuous metric, so that the score-based attack techniques can be adopted. \cite{dong2019efficient} adopts evolution algorithm to achieve the decision-based attack against face recognition system.
% \Xiaojun{remove redundancy in the background part; mention other gradient estimation on BA} HopSkipJumpAttack~\cite{chen2019hopskipjumpattack} improves upon the basic random walk idea in the original Boundary Attack paper. It proposes an unbiased estimate of the gradient of the model on the decision boundary via a Monte-Carlo algorithm. The \queryimages are generated by adding the \boundaryimage with a set of noise vectors that are randomly sampled from the whole image space. Using the estimated gradient, the paper is able to perform more efficient update to get to a \boundaryimage that is closer to the \targetimage by taking a step towards the gradient direction and then performing binary search back to the decision boundary iteratively. This method reduces the query number compared with Boundary Attack, but since it is sampling from an extremely high-dimensional space(for example, ImageNet data samples lie in $224\times 224\times 3$ dimensional space), the number of queries required to get a fair estimation for updating is still large.
\vspace{-5mm}
\paragraph{Dimension Reduction in Score-based Attack}
Another line of work involves the dimension reduction techniques only for the score-based attacks, which requires access to the prediction of confidence for each class.
In~\cite{guo2018low}, the authors draw intuition from JPEG codec~\cite{wallace1992jpeg} image compression techniques and propose to use discrete cosine transform (DCT) for generating low frequency adversarial perturbations to assist score-based adversarial attack.
AutoZoom~\cite{tu2019autozoom} trains an auto-encoder offline with natural images and uses the decoder network as a dimension reduction tool. Constrained perturbations in the latent space of the auto-encoder are generated and passed through the decoder. The resulting perturbation in the image space is added to the benign one to obtain a query sample. 

% \section{Discussion and conclusions}
% % talk about potential defense

\section{Conclusion}
Overall we propose \name, a general query-efficient boundary-based blackbox attack framework. We in addition explore three novel subspace optimization approaches to reduce the number of queries from spatial, frequency, and intrinsic components perspectives. 
Based on our theoretic analysis, we show the optimality of the proposed subspace based gradient estimation compared with the estimation over the original space. 
Extensive results show that the proposed \name significantly reduces the required number of queries and yields high quality adversarial examples against both offline and onlie real-world APIs.

\subsection*{Acknowledgement}
We would like to thank Prof. Yuan Qi and Prof. Le Song for their comments and advice in this project. This work is partially supported by NSF grant No.1910100.

\newpage

{\small
\bibliographystyle{ieee_fullname}
\bibliography{bibliography}
}

\clearpage
\appendix
% \section{Why Direct Gradient Estimation is Inefficient}
% Eqn.\ref{eqn:grad-est} provides a way to estimate gradients via decision-based black-box access to the model. However, this estimation may not be accurate enough by noticing that $\widetilde{\nabla S}$ is a linear combination of $B$ vectors in $\mathbb{R}^m$. In practice, $m$ may be very large (e.g. $3\times224\times224$ in most pretrained models on ImageNet) while $B$ is chosen to be small (e.g. 100). Therefore, even in the best case the approximation is only the projection of the gradient $\nabla S$ onto the subspace spanned by $u_1,\ldots,u_B$. Considering that $u$'s are randomly chosen, the approximation may not be good. In particular, we have the following theorem on the expected quality of the estimated gradient:

\section{Proof of Theorem \ref{tho:dimred}}
\label{sec:tho-proof}
We first prove a lemma of the gradient estimation quality which samples from the entire subspace:
% The gradient vector estimated by Eqn. \ref{eqn:grad-est} is a linear combination of $B$ vectors $u_1,\ldots,u_B \in \mathbb{R}^m$.  The best estimated gradient we can possibly get is the projection of the actual gradient onto the subspace spanned by the $B$ vectors. In practice, $m$ is large (for example, $3\times224\times224$ on ImageNet) while $B$ is chosen to be small (e.g., 100). The following theorem shows the expectation of the gradient approximation in a randomly sampled subspace. 
\begin{lemma}
\label{lemma:cos}
For a boundary point $x$, suppose that $S(x)$ has $L$-Lipschitz gradients in a neighborhood of $x$, and that ${\bf u}_1, \ldots, {\bf u}_B$ are sampled from the unit ball in $\mathbb{R}^m$ and orthogonal to each other. Then the expected cosine similarity between $\widetilde{\nabla S}$ and $\nabla S$ can be bounded by:
\begin{align}
    & \bigg( 2\bigg(1-(\frac{L\delta}{2||\nabla S||_2})^2\bigg)^{\frac{m-1}{2}} - 1 \bigg)c_m\sqrt{\frac B m}\\
    \leq & \mathbb{E}\big[\cos (\widetilde{\nabla S}, \nabla S) \big]\\
    \leq & c_m\sqrt{\frac B m}
\end{align}
% \begin{align}
%     \bigg( 2\bigg(1-(\frac{L\delta}{2||\nabla S||_2})^2\bigg)^{\frac{m-1}{2}} - 1 \bigg)c_m\sqrt{\frac B m}
%     & \leq \mathbb{E}\big[\cos (\widetilde{\nabla S}, \nabla S) \big]\\
%     \mathbb{E}\big[\cos (\widetilde{\nabla S}, \nabla S) \big] &\leq c_m\sqrt{\frac B m}
% \end{align}
where $c_m$ is a constant related with $m$ and can be bounded by $c_m \in (2/\pi, 1)$. In particular, we have:
\begin{align}
    \label{eqn:grad-est-qual}
    \lim_{\delta\rightarrow 0}\mathbb{E}\big[\cos (\widetilde{\nabla S}, \nabla S) \big] = c_m\sqrt{\frac B m}.
\end{align}
\end{lemma}

\label{sec:proof}
\begin{proof}
Let ${\bf u}_1, \ldots, {\bf u}_B$ be the random orthonormal vectors sampled from $\mathbb{R}^m$. We expand the vectors to an orthonormal basis in $\mathbb{R}^m$: ${\bf q}_1={\bf u}_1, \ldots, {\bf q}_B={\bf u}_B, {\bf q}_{B+1}, \ldots, {\bf q}_m$. Hence, the gradient direction can be written as:
\begin{equation}
    \label{eqn:proof-gt}
    \frac{\nabla S}{||\nabla S||_2} = \sum_{i=1}^m a_i{\bf q}_i
\end{equation}
where $a_i=\langle \frac{\nabla S}{||\nabla S||_2}, {\bf q}_i \rangle$ and its distribution is equivalent to the distribution of one coordinate of an $(m-1)$-sphere. Then each $a_i$ follows the probability distribution function:
\begin{equation}
    p_a(x) = \frac{(1-x^2)^{\frac{m-3}2}}{\mathcal{B}(\frac{m-1}2, \frac12)},~x\in(-1,1)
\end{equation}
where $\mathcal{B}$ is the beta function. According to the conclusion in the proof of Theorem 1 in \cite{chen2019hopskipjumpattack}, if we let $w=\frac{L\delta}{2||\nabla S||_2}$, then it always holds true that $\phi({\bf x}+\delta{\bf u_i})=1$ when $a_i>w$, -1 when $a_i<-w$ regardless of $u_i$ and the decision boundary shape. Hence, we can rewrite $\phi_i$ in term of $a_i$:
\begin{equation}
    \phi_i = \phi({\bf x}+\delta{\bf u_i}) =
    \begin{cases}
    1, & \text{if } a_i \in [w, 1)\\
    -1, & \text{if } a_i \in (-1, -w]\\
    \text{undetermined}, & \text{otherwise}
    \end{cases}
\end{equation}
Therefore, the estimated gradient can be rewritten as:
\begin{equation}
    \label{eqn:proof-est}
    \widetilde{\nabla S}=\frac1B\sum_{i=1}^B \phi_i{\bf u}_i
\end{equation}
Combining Eqn. \ref{eqn:proof-gt} and \ref{eqn:proof-est}, we can calculate the cosine similarity:
\begin{align}
    \mathop{\mathbb{E}}\big[ \cos (\widetilde{\nabla S}, \nabla S)\big] &= \mathop{\mathbb{E}}_{a_1, \ldots,a_B}\frac{\sum_{i=1}^Ba_i\phi_i}{\sqrt{B}}\\
    &= \sqrt{B}\cdot \mathop{\mathbb{E}}_{a_1} \big[ a_1\phi_1 \big]
\end{align}
In the best case, $\phi_1$ has the same sign with $a_1$ everywhere on $(-1,1)$; in the worst case, $\phi_1$ has different sign with $a_1$ on $(-w,w)$. In addition, $p_a(x)$ is symmetric on $(-1, 1)$. Therefore, the expectation is bounded by:
\begin{align}
    &2\int_{w}^{1} p_a(x)\cdot xdx - 2\int_0^w p_a(x)\cdot xdx\\
    \leq& \mathop{\mathbb{E}}_{a_1} \big[ a_1\phi_1 \big]\\
    \leq& 2\int_0^1 p_a(x)\cdot xdx
\end{align}
By calculating the integration, we have:
\begin{align}
&\bigg( 2\bigg(1-w^2\bigg)^{\frac{m-1}{2}} - 1 \bigg) \cdot \frac{2\sqrt{B}}{\mathcal{B}(\frac{m-1}2, \frac12)\cdot(m-1)}\\
\leq& \mathop{\mathbb{E}}\big[ \cos (\widetilde{\nabla S}, \nabla S)\big]\\
\leq& \frac{2\sqrt{B}}{\mathcal{B}(\frac{m-1}2, \frac12)\cdot(m-1)}
\end{align}
The only problem is to calculate $\mathcal{B}(\frac{m-1}2, \frac12)\cdot(m-1)$. It is easy to prove by scaling that $\mathcal{B}(\frac{m-1}2, \frac12)\cdot(m-1) \in (2\sqrt{m}, \pi \sqrt{m})$. Hence we can get the conclusion in the theorem.
\end{proof}

Having Lemma \ref{lemma:cos}, Theorem \ref{tho:dimred} follows by noticing that $\mathbb{E}\big[\cos (\widetilde{\nabla S}, \nabla S) \big]=\rho\mathbb{E}\big[\cos (\widetilde{\nabla S}, \text{proj}_{\text{span}(W)}(\nabla S)) \big]$.

\end{document}